\documentclass[11pt,a4paper]{article}
\usepackage[hyperref]{conll-2019}
\usepackage{times}
\usepackage{latexsym}
\usepackage{url}

\usepackage{graphicx}
\usepackage{float}
\usepackage{amsmath}
\usepackage{amssymb}
\usepackage{multirow}
\usepackage{booktabs}
\usepackage{array}
\usepackage{algorithm}
\usepackage{algorithmic}
\usepackage{longtable}
\usepackage{enumitem}
\usepackage{bm}
\usepackage{subfigure}
\usepackage{color, soul}
\usepackage{caption}
\usepackage[normalem]{ulem}
\usepackage{url}

\aclfinalcopy % Uncomment this line for the final submission

\title{TripleNet: Triple Attention Network for Multi-Turn Response Selection in Retrieval-based Chatbots }
  
\author{
Wentao Ma$^\dag$,
Yiming Cui$^\ddag$$^\dag$,
Nan Shao$^\dag$,  \\
\textbf{Su He$^\dag$}, 
\textbf{Wei-Nan Zhang$^\ddag$},
\textbf{Ting Liu$^\ddag$},
\textbf{Shijin Wang$^\dag$$^\S$},
\textbf{Guoping Hu$^\dag$}\\
{$^\dag$State Key Laboratory of Cognitive Intelligence, iFLYTEK Research, China} \\
{$^\ddag$Research Center for Social Computing and Information Retrieval (SCIR), } \\
{Harbin Institute of Technology, Harbin, China} \\
{$^\S$iFLYTEK AI Research (Hebei), Langfang, China} \\
$^\dag$$^\S$\tt\{wtma,ymcui,nanshao,suhe,sjwang3,gphu\}@iflytek.com \\
$^\ddag$\tt\{ymcui,wnzhang,tliu\}@ir.hit.edu.cn
}

\date{}

\begin{document}
\maketitle
\begin{abstract}
 We consider the importance of different utterances in the context for selecting the response usually depends on the current query.\footnote {In this paper, we define the last message which is waiting for a response as the `query,'  the conversation history including the query  as `context,' and  each message in the context as an `utterance.'}
 In this paper, we propose the model TripleNet to fully model the task with the triple $ \left\langle{context, query, response} \right\rangle$  instead of $ \left\langle context, response \right\rangle$ in previous works. 
The heart of TripleNet is a novel attention mechanism named triple attention to model the relationships within the triple at four levels. 
 The new mechanism updates the representation for each element based on the attention with the other two concurrently and symmetrically.
 We match the triple $ \left\langle{C, Q, R} \right\rangle$ centered on the response from char to context level for prediction.
 Experimental results on two large-scale multi-turn response selection datasets show that the proposed model can significantly outperform the state-of-the-art methods. \footnote{ TripleNet source code is available at \url{https://github.com/wtma/TripleNet}.}
 \end{abstract}

\section{Introduction}\label{introduction}
To establish a human-machine dialogue system is one of the most challenging tasks in Artificial Intelligence (AI). 
Existing works on building dialogue systems are mainly divided into two categories: retrieval-based method \cite{Yan-2016-Learning,Zhou-2016-Multi}, and generation-based method \cite{Vinyals-2015A}.
The retrieval-based method retrieves multiple candidate responses from the massive repository and selects the best one as the system's response, while the generation-based method uses the encoder-decoder framework to generate the response, which is similar to machine translation. 

 \begin{figure}[tp]
\centering   \small   
 \begin{tabular}{m{7cm}}
 \toprule
\textbf {A}:  i downloaded angry ip scanner and now it doesn't work and \uline{i can't \textbf {uninstall} it }\\
\textbf {B}:  you \uline{\textbf {installed}} it via package or via some  \uwave{binary installer}\\
\textbf {A}:  \uline{i \textbf {installed}} from ubuntu soft center\\ 
\textbf {B}:  hm i do n't know what package it is but it should let you remove it the same way \\
\textbf {A}: ah makes sense then ... hm \uwave{was it a deb \textbf {file} }\\
\midrule
\textbf {True Response}: i think \uwave{it was another \textbf {format}} mayge sth starting with r \\
\textbf {False Response}:  thanks i appreciate it try \uline{sudo apt-get \textbf {install}} libxine-extracodecs \\
 \bottomrule
 \end{tabular}
 \caption{\label{case-ubuntu} A real example in the Ubuntu Corpus. The upper part is the conversation between speaker A and B. The speaker A want to uninstall the ip scanner and the current query is about the format of the package, so the true response is about the format, but the existing conversation model can be easily misled by the high frequency term `install' as they deal with the query and other utterances in the same way.}
 \end{figure}
In this paper, we are focusing on the retrieval-based method because it is more practical in applications.
Selecting a response from a set of candidates is an important and challenging task for the retrieval-based method.
Many of the previous approaches are based on Deep Neural Network (DNN) to select the response for single-turn conversation \cite{Lu-2013-A}.
We study multi-turn response selection in this paper, which is rather difficult because it not only requires identification of the important information such as keywords, phrases, and sentences, but also the latent dependencies between the context,  query,  and candidate response. 
 
 Previous works \cite{zhou-2018-multi,Wu-2017-Sequential} show that representing the context at different granularities is vital for multi-turn response selection.
 However, it is not enough for multi-turn response selection.  Figure \ref{case-ubuntu} illustrates the problem with a real example in Ubuntu Corpus.  
As demonstrated, the following two points should be modeled to solve the problem:
 (1) the importance of current query should be highlighted, because it has great impact on the importance of different utterances in the context.
 For example, the query in the case is about the format of the file (`deb file'), which leads the last two utterances (including the query) are more important than the previous ones. If we only match the response with the context, the model may be misled by the high frequency word `install' and choose the false candidate.
 (2) the information of different granularities is important, which includes not only the word, utterance, and context level, but also the char level. For example, the different tenses (`install,' `installed') and the misspelling word (`angry') appear constantly in the conversation.  
 Similar to the role of question for the task of machine reading comprehension \cite{seo-2016-bidirectional, cui-acl2017-aoa, chen-2019-convolutional}, the query in this task is also the key to selecting the response.
In this paper, we propose a model named TripleNet to excavate the role of query. The main contributions of our work are listed as follows.
 \begin {itemize}
 \item we use a novel triple attention mechanism to model the relationships within $ \left\langle{C, Q, R} \right\rangle$ instead of $ \left\langle{C, R} \right\rangle$; 
 \item we propose a hierarchical representation module to fully model the conversation from char to context level;  
 \item The experimental results on Ubuntu and Douban corpus show that TripleNet significantly outperform the state-of-the-art result.
 \end {itemize}

%%%%%%%%%%%%%%%%%%%%%%%%%%%%%%%%%%%%%%%%%
\section{Related Works}\label{related-works}
Earlier works on building the conversation systems are generally based on rules or templates \cite {Walker-2001-Quantitative}, which are designed for the specific domain and need much human effort to collect the rules and domain knowledge. 
As the portability and coverage of such systems are far from satisfaction, people pay more attention to the data-driven approaches for the 
open-domain conversation system \cite{Ritter-2011-Data, Higashinaka-2014-Towards}. 
The main challenge for open-domain conversation is to produce a corresponding response based on the current context. 
As mentioned previously, the retrieval-based and generation-based methods are the mainstream approaches for conversational response generation. In this paper, we focus on the task response selection which belongs to retrieval-based approach.

The early studies of response selection generally focus on the single-turn conversation, which use only the current query to select the response \cite{Lu-2013-A, Ji-2014-An,Li-2015-Syntax}.
Since it is hard to get the topic and intention of the conversation by single-turn, researchers turn their attention to multi-turn conversation and model the context instead of the current query to predict the response. 
First, \citet{Lowe-2015-The} released the Ubuntu Dialogue dataset and proposed a neural model which matches the context and response with corresponding representations via RNNs and LSTMs. 
\citet{Kadlec-2015-Improved} evaluate the performances of various models on the dataset, such as LSTMs, Bi-LSTMs, and CNNs. 
Later, \citet{Yan-2016-Learning} concatenated utterances with the reformulated query and various features in a deep neural network. 
\citet{Baudi-2016-Sentence} regarded the task as sentence pair scoring and implemented an RNN-CNN neural network model with attention. 
\citet{Zhou-2016-Multi} proposed a multi-view model with CNN and RNN, modeling the context in both word and utterance view. 
Further, \citet{Xu-2017-Incorporating} proposed a deep neural network to incorporate background knowledge for conversation by LSTM with  a specially designed recall gate. 
\citet{Wu-2017-Sequential} proposed matching the context and response by their word and phrase representations, which had significant improvement from previous work. 
\citet{zhang-2018-modeling} introduced a self-matching attention to route the vital information in each utterance, and used RNN to fuse the matching result. 
\citet{zhou-2018-multi} used self-attention and cross-attention to construct the representations at different granularities, achieving a state-of-the-art result.

Our model is different from the previous methods: first we model the task with the triple $ \left\langle{C, Q, R} \right\rangle$  instead of $ \left\langle C, R \right\rangle$ in the early works, and use a novel triple attention matching mechanism to model the relationships within the triple.  Then we represent the context from low (character) to high (context) level,  which constructs the representations for the context more comprehensively. 
\begin {figure*} [t]
  \centering
  \includegraphics [width= 0.9\textwidth] {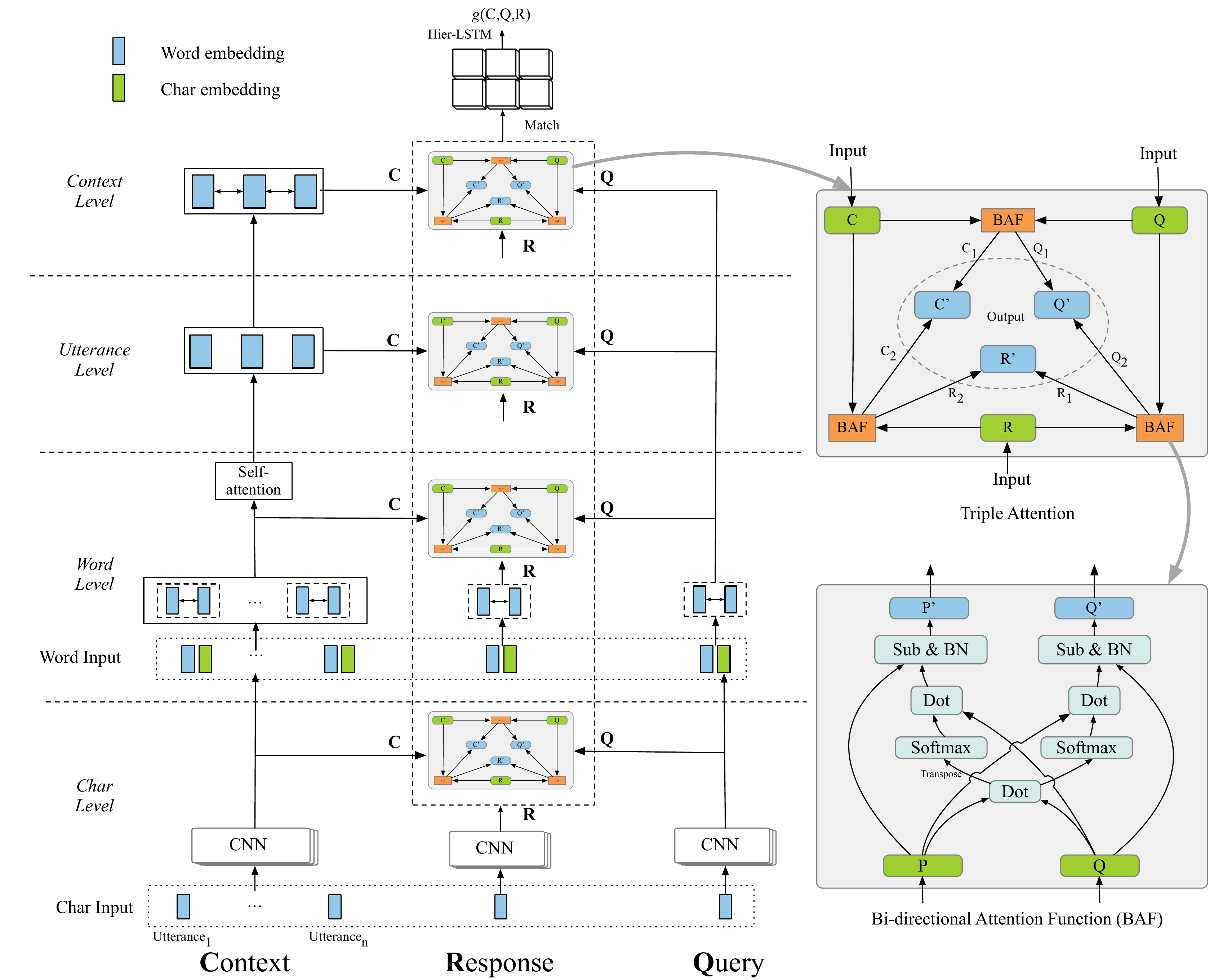}
  \caption{\label{HAN-arch} The neural architecture of the model TripleNet. \emph{(best viewed in color)}}
\end {figure*}

%%%%%%%%%%%%%%%%%%%%%%
\section{Model}\label{Model}
In this section, we will give a detailed introduction of the proposed model TripleNet.
We first formalize the problem of the response selection for multi-turn conversation. 
Then we briefly introduce the overall architecture of the proposed model. 
Finally, the details of each part of our model will be illustrated.

\subsection{Task Definition}
For the response selection, we define the task as given the context $C$, current query $Q$ and candidate response $R$, which is different  from almost all the previous works  \cite{zhou-2018-multi,Wu-2017-Sequential}. We aim to build a model function $g(C, Q, R)$ to predict the possibility of the candidate response to be the correct response.
\begin {equation}
score = g(C, Q, R)
\end{equation}
The information in context is composed of four levels: context, utterances, words and characters, which can be formulated as $C = (u_{1}, u_{2}, ... , u_{i}, ..., u_{n})$, where $u_i$ represents the $i$th utterance, and $n$ is the maximum utterance number. 
The last utterance in the context is query $Q=U_n$; we still use query as the end of context to maintain the integrity of the information in context.
Each utterance can be formulated as $u_i = (w_1, ..., w_j, .., w_{m})$, where $w_j$ is the $j$th word in the utterance and $m$ is the maximum word number in the utterance.
Each word can be represented by multiply characters $w_j = (ch_1, ..., ch_k, .., ch_{l})$, where $ch_k$ is the $k$th char and $l$ is the length of the word in char-level.  The latter two levels are similar in the query and response.

\subsection{Model Overview}
The overall architecture of the model TripleNet is displayed in Figure \ref {HAN-arch}.  
The model has a bottom-up architecture that organizes the calculation from char to context level. 
In each level, we first uses the hierarchical representation module to construct the representations of context, response and query. 
Then the triple attention mechanism is applied to update the representations. At last, the model matches them while focused on the response and fuses the result for prediction.

In the hierarchical representation module, we represent the conversation in four perspectives including char, word, utterance, and context.
In the char-level, a convolutional neural network (CNN) is applied to the embedding matrix of each word and produces the embedding of the word by convolution and maxpooling operations as the char-level representation.
In word-level, we use a shared LSTM layer to obtain the word-level embedding for each word. 
After that, we use self-attention to encode the representation of each utterance into a vector which is the utterance-level representation.
At last, the utterance-level representation of each utterance is fed into another LSTM layer to further model the information among different utterances, forming the context-level representation.

 The structure of the triple attention mechanism can be seen in the right part of Figure \ref {HAN-arch}. We first design a bi-directional attention function (BAF) to calculate the attention between two sequences and output their new representations.
To model the relationship of the triple  $\left\langle{C, Q, R} \right\rangle$, we apply BAF to each pair within the triple and get two new representations for each one element, and then we add them together as its final attention-based representation.
In the triple attention mechanism, we can update the representation of each one based on the attention result  with the other two simultaneously, and each element participates in the whole calculation in the same way.

%%%%%%%%%%%%%%%%%%%%%%
\subsection{Hierarchical Representation}
{\noindent{\bf Char-level Representation.}} At first, we embed the characters in each word into fixed size vectors and use a CNN followed by max-pooling to get character-derived embeddings for each word, which can be formulated by
\begin {gather}
ch_{j,t} = tanh(W_1^j * x_{t:t+s_j-1} + b_1^j) \\
ch_j =  MaxPooling_{t=0}^L[ch_{j,t} ]
\end {gather}
where $W_1^j$, $b_1^j$ are parameters, $x_{t:t+s_j-1}$ refers to the concatenation of the embedding of ($x_t$,...,$x_{t+s_j-1}$), $s_j$ is the window size of $j$th filter, and the $ch$ is the representation of the word in char-level.

{\noindent{\bf Word-level Representation.}} Furthermore, we embed word $x$ by pre-trained word vectors, and we also introduce a word matching (MF) feature to the embedding to make the model more sensitive to concurrent words. If the word appears in the response and context or query simultaneously, we set the feature to 1, otherwise to 0. 
\begin {gather}
e(x) = [W_{e} \cdot x; ch(x); MF]
\end {gather}
where  $e(x)$ to denotes the embedding representation, $W_{e}$ is the pre-trained word embedding,  and $ch(x)$ is the character embedding function. 
We use a shared bi-directional LSTM to get contextual word representations in each utterance, query, and the response. 
The representation of each word is formed by concatenating the forward and backward LSTM hidden output.
\begin {gather}
\overleftarrow {h(x)} = \overleftarrow{\text{LSTM}}(e(x)) \\
{\overrightarrow{h(x)}} = \overrightarrow{\text{LSTM}}(e(x)) \\
h(x) = [\overleftarrow {h(x)};\overrightarrow {h(x)} ]
\end {gather}
where $h(x)$ is the representation of the word. We denote the word-level representation of the context as $h_{u} \in \mathbb{R} ^ { m * d_w} $ and the response as $h_{r} \in \mathbb{R} ^ {m * d_w} $, where $d_w$ is the dimension of Bi-LSTMs. Until now, we have constructed the representations of context, query, and  response in char and word level, and  we only represent the latter two in these two levels because they don't  have such rich contextual information as the context.

{\noindent{\bf Utterance-level Representation.}}
Given the $k_{th}$ utterance ${u_k} = [h^i_{u_k}]^m_{i=1}$, we construct the utterance-level representation by self-attention \cite{lin-2017-structured}:
\begin {gather}
\alpha_i^k =  softmax(W_{3}  tanh(W_{2}h_{u_k}(i)^T)) \\
u_{k} = \sum\nolimits_{i=1}^m h^i_{u_k} \alpha_i^k
\end {gather}
where $W_2 \in \mathbb{R}^{d * d_w}$,  $W_3 \in \mathbb{R}^{d}$ are trainable weights,  $d$ is a hyperparameter, $u_{k}$ is the utterance-level representation,  and $\alpha_i^k$ is the attention weight for the $i$th word in the $k$th utterance, which signifies the importance of the word in the utterance. 

{\noindent{\bf Context-level Representation.}}
To further model the continuity and contextual information among the utterances, we fed the utterance-level representations into another bi-directional LSTM layer to obtain the representation for each utterance in context perspective. 
\begin {gather}
c_{k} = \text{Bi-LSTM}([u_{k}]_{k=1}^n)
\end {gather}
where $c_{k} \in \mathbb{R}^{d_c}$ is the context-level representation for the $k$th utterance in the context and $d_c$ is the output size of the Bi-LSTM.

\subsection{Triple Attention}
In this part, we update the representations of context, query, and response in each level by triple attention, the motivation of which is to model the latent relationships within $ \left\langle{context, query, response} \right\rangle$ .

Given the triple $ \left\langle{C, Q, R} \right\rangle$ , we fed each of its pairs into bi-directional attention function (BAF).
 \begin {gather}
C_1, Q_1 = BAF(C, Q) \\
C_2, R_1 = BAF(C, Q) \\
Q_2, R_2 = BAF(C, R)  \\
C' = BN(C_1 + C_2) \\
Q' = BN(Q_1+Q_2) \\
R' = BN(R_1+R_2)
\end {gather}
where $BN$ denotes the batch normalization layer \cite{ioffe-2015-batch} which is conducive to  preventing vanishing or exploding of gradients.
 $BAF$ produces the new representations for two sequences (P, Q) by the attention from two directions, which is inspired by \citet{seo-2016-bidirectional}.  We can formulate it by 
 \begin {gather}
 M_{pq} = P^T tanh(W_3 Q) \\
 Att_{pq} =  softmax( M_{pq}) \\
 Att_{qp} = softmax( M_{pq}^T) \\
P' = P - \tilde{Q}; ~~ \tilde{Q} = Q  Att_{pq}; \\
Q' = Q - \tilde{P}; ~~ \tilde{P} = P  Att_{qp};
\end {gather}
where $Att_{pq}$, $Att_{qp}$ are the attention between $P$ and $Q$ in two directions,  $P'$, $Q'$ are the new representations the two sequences (P, Q), and we apply a batch normalization layer upon them too.

We find that the triple attention has some interesting features: (1) triple, the representation for each element in the triple $\left\langle{C, Q, R} \right\rangle$ is updated based on the attention to the other two concurrently; (2) symmetrical, which means each element in the triple plays the same role in the structure because their contents are similar in the whole conversation; (3) unchanged dimension, all the outputs of triple attention has the same dimensions as the inputs, so we can stack multiple layers as needed.

\subsection{Triple Matching and Prediction}
{\noindent{\bf Triple Matching.}} We match the triple $\left\langle{C,Q,R} \right\rangle$ in each level with the cosine distance using new representations produced by triple attention. This process focuses on the response because it is our target. For example, in the char-level, we match the triple by
\begin {gather}
\tilde M_{rc}^1 (i, k, j) = cosine(ch_{r}'(i), ch_{u_k}'(j)) \\
M_{rc}^1(i, k) =  \max \limits_{0<j<m}  \tilde M_1(i, j, k) \\
M_{rq}^1 (i, j) =  cosine(ch_{r}'(i), ch_{q}'(j)) \\
M_1 = [M_{rc}^1(i, k);M_{rq}^1 (i, j)]
\end {gather}
where $ch'$ is the representation updated by triple attention, $M_1 \in \mathbb{R}^{m * (n+m)}$ is the char-level matching result,  the word-level matches the triple in the same way, and the utterance and the context level match the triple without the maxpooling operation. We use $M_2$, $M_3$, $M_4$  as the matching results in the word, utterance and context levels.

{\noindent{\bf Fusion.}}
After obtaining the four-level matching matrix, we use hierarchical RNN to get highly abstract features.
Firstly, we concatenate the four matrices to form a 3D cube $M \in \mathbb{R}^{m*(n+m)*4}$ and we use $m$ as one of the matrix in $M$, which denotes the matching result for one word in response in four levels. 
\begin {gather}
M = [M_1;M_2;M_3;M_4] \\
\tilde{m} = MaxPooling_{i=0}^{n+m}[\text{Bi-LSTM}(m_i)]\\
v = MaxPooling_{j=0}^m[\text{Bi-LSTM}{(\tilde{m}_j})]
\end {gather}
Where $m_i$ and $\tilde{m}_j$ are the $i$th, $j$th row in the matrix $m$ and $\tilde{m}$. We merge the results from different time steps in the outputs of LSTM by max-pooling operation. Until now, we encode the matching result into a single feature vector $v$.

{\noindent{\bf Final Prediction.}}
For the final prediction, we fed the vector $V$ into a full-connected layer with sigmoid output activation.
\begin {equation}
g(C, Q, R) =  sigmoid(W_4 \cdot v+b_4) \\
\end {equation}
where $W_4, b_4$ are trainable weights.
Our purpose is to predict the matching score between the context, query and candidate response, which can be seen as a binary classification task. To train our model, we minimize the cross entropy loss between the prediction and ground truth.

%%%%%%%%%%%%%%%%%%%%%%%%%%%%%%%%%%%%%%%%%
\section{Experiments}\label{experiments}        
 \begin{table*}[t!]
        \begin{center}\small
        \begin{tabular}{l cccc | cccccc}
        \toprule
        &  \multicolumn{4}{c}{\bf {Ubuntu Dialogue Corpus}} & \multicolumn{6}{c}{\bf {Douban Conversation Corpus} } \\
        & R${_2}$@1& R$_{10}$@1& R$_{10}$@2 & R$_{10}$@5 & MAP & MRR & P@1 & R$_{10}$@1 & R$_{10}$@2 &R$_{10}$@5 \\ 
        \midrule
        DualEncoder & 90.1 & 63.8 & 78.4 & 94.9 & 48.5 & 52.7 & 32.0 & 18.7 & 34.3 & 72.0 \\
        MV-LSTM & 90.6 & 65.3 & 80.4 & 94.6 & 49.8 & 53.8 & 34.8 & 20.2 & 35.1 & 71.6 \\
        Match-LSTM & 90.4 & 65.3  & 80.4  & 94.6 & 49.8 & 53.8 & 34.8 & 20.2 & 34.8 & 71.0 \\
        DL2R & 89.9 & 62.6  & 78.3  & 94.4 & 48.8 & 52.7 & 33.0 & 19.3 & 34.2 & 70.5 \\
        Multi-View &  90.8 & 66.2 & 80.1 & 95.1 & 50.5 & 54.3 & 34.2 & 20.2 & 35.0 & 72.9\\
        SMN & 92.6 & 72.6 & 84.7 & 96.1 & 52.9 & 56.9 & 39.7 & 23.3 & 39.6 & 72.4 \\
        \midrule
        RNN-CNN & 91.1 & 67.2 & 80.9 & 95.6 & - & - & - & - & - & - \\
        DUA & - & 75.2 & 86.8 & 96.2 & \emph{55.1} & 59.9 & 42.1 & 24.3 & \emph{42.1} &  \emph {78.0} \\
        DAM & \emph{93.8} & \emph{76.7} & \emph{87.4} & \emph{96.9} & 55.0 & \emph{60.1} & \emph{42.7} & \emph{25.4} & 41.0 & 75.7 \\
        \midrule
        \midrule
        TripleNet &\bf 94.3 &\bf 79.0 &\bf 88.5 &\bf 97.0 &\bf 56.4 &\bf 61.8 &\bf 44.7 &\bf 26.8 &\bf 42.6 & 77.8 \\
        TripleNet$_{elmo}$ & 95.1 & 80.5 & 89.7 & 97.6 & 60.9  & 65.0 & 47.0 & 27.8 & 48.7 & 81.4 \\
        TripleNet$_{ensemble}$ & 95.6 & 82.1 & 90.9 & 98.0 & 63.2 & 67.8 & 51.5 & 31.3 & 49.4 & 83.2 \\
        \bottomrule
        \end{tabular}
        \end{center}
        \caption{\label{result-chat} Experimental results on two public dialogue datasets. The table is segmented into three sections:  Non-Attention models, Attention-based models and our models. The italics denotes the previous best results, and the scores in bold express the new state-of-the-art result of single model without any pre-training layer.}
        \end{table*}

\subsection{Dataset}
We first evaluate our model on Ubuntu Dialogue Corpus \cite {Lowe-2015-The} because it is the largest public multi-turn dialogue corpus which consists of about one million conversations in the specific domain.
To reduce the number of unknown words, we use the shared copy of the Ubuntu corpus by \citet{Xu-2017-Incorporating} which replaces the numbers, paths, and URLs by specific symbols.\footnote {\url{https://www.dropbox.com/s/ 2fdn26rj6h9bpvl/ubuntudata.zip}}
Furthermore, to verify the generalization of our model, we also carry out experiments on Douban Conversation Corpus \cite {Wu-2017-Sequential}, which shares similar format with the Ubuntu corpus but is open-domain and in the Chinese language.

For the Ubuntu corpus, we use the recall at position k in $n$ candidate responses ($R_n@k$) as evaluation metrics, and we use MAP (Mean Average Precision), MRR (Mean Reciprocal Rank), and Precision-at-one as the additional metrics for Douban corpus, following the previous work \cite {Wu-2017-Sequential}.

\subsection{Experiment Setup}
We implement our model by Keras \cite {chollet-2015-keras} with TensorFlow backend. 
In the Embedding Layer, the word embeddings are pre-trained using the training set via GloVe \cite{pennington-etal-2014}, the weights of which are trainable. For char embedding, we set the kernel shape as 3 and filter number as 200 in the CNN layer. 
 For all the Bi-directional LSTM layers, we set  their hidden size to 200.
We use Adamax \cite{kingma-2014-adam} for weight updating with an initial learning rate of 0.002.
For ensemble models, we generate 6 models for each corpus using different random seeds and merge the result by voting.

For better comparison with the baseline models, the main super parameters in TripleNet, such as the embedding size, max length of each turn, and the vocabularies,  are the same as those of the baseline models. 
The maximum number of conversation turns, which changes with the models, is 12 in our model,  9 in DAM \cite {Wu-2017-Sequential}, and 10 in SMN \cite {Wu-2017-Sequential}.

\subsection{Baseline Models}
We basically divided baseline models into two categories for comparisons.
\\ \textbf {Non-Attention Models.} The majority of the previous works on this task are designed without attention mechanisms, including the Sequential Matching Network (SMN) \cite {Wu-2017-Sequential}, Multi-View model \cite{Zhou-2016-Multi}, Deep Learning to Respond (DL2R) \cite{Yan-2016-Learning}, Match-LSTM \cite {wang-2016-learning}, MV-LSTM \cite {wan-2016-match}, and DualEncoder  \cite {Lowe-2015-The}.
\\ \textbf {Attention-based Models.} The attention-based models typically match the context and the candidate response based on the attention among them, including DAM \cite {zhou-2018-multi}, DUA \cite  {zhang-2018-modeling},  and RNN-CNN \cite {Baudi-2016-Sentence}.

\subsection{Overall Results}
The overall results on two datasets are depicted in Table \ref {result-chat}.
Our results are obviously better on the two datasets compared with recently attention-based model DAM, which exceeds 2.3\% in $R_{10}$@1 of Ubuntu and 2.6\% in $P@1$ of Douban. 
Furthermore, our score is significantly exceeding in almost all metrics except the $R_{10}$@5 in Douban when compared with DUA, which may be because the metric is not very stable as the test set in Douban is very small (1000).

To further improve the performance, we utilize pre-trained ELMo \cite{peters-elmo-2018} and fine-tune it on the training set in the Ubuntu condition while we train ELMo from scratch using the Douban training set.
As the baseline of Douban corpus is relatively lower, we observe much bigger improvements in the corpus using ELMo.
The model ensemble has further improvements based on the single model with ELMo;  the score of $R_{10}$@1 in Ubuntu is close to the average performance of human experts at 83.8 \cite{lowe-2016-evaluation}.

Compared to non-attention models such as the SMN and Multi-view, which match the context and response at two levels,  TripleNet shows substantial improvements. On $R_{10}$@1 for Ubuntu corpus, there is  a 6.3\% absolute improvement from SMN and 12.8\% from Multi-view, showing the effectiveness of triple attention.

%\vspace{0.2 pt}
\begin{figure*}[htbp] 
\centering
\subfigure{
\begin{minipage}[c]{0.31\linewidth}
\includegraphics[width=1\linewidth]{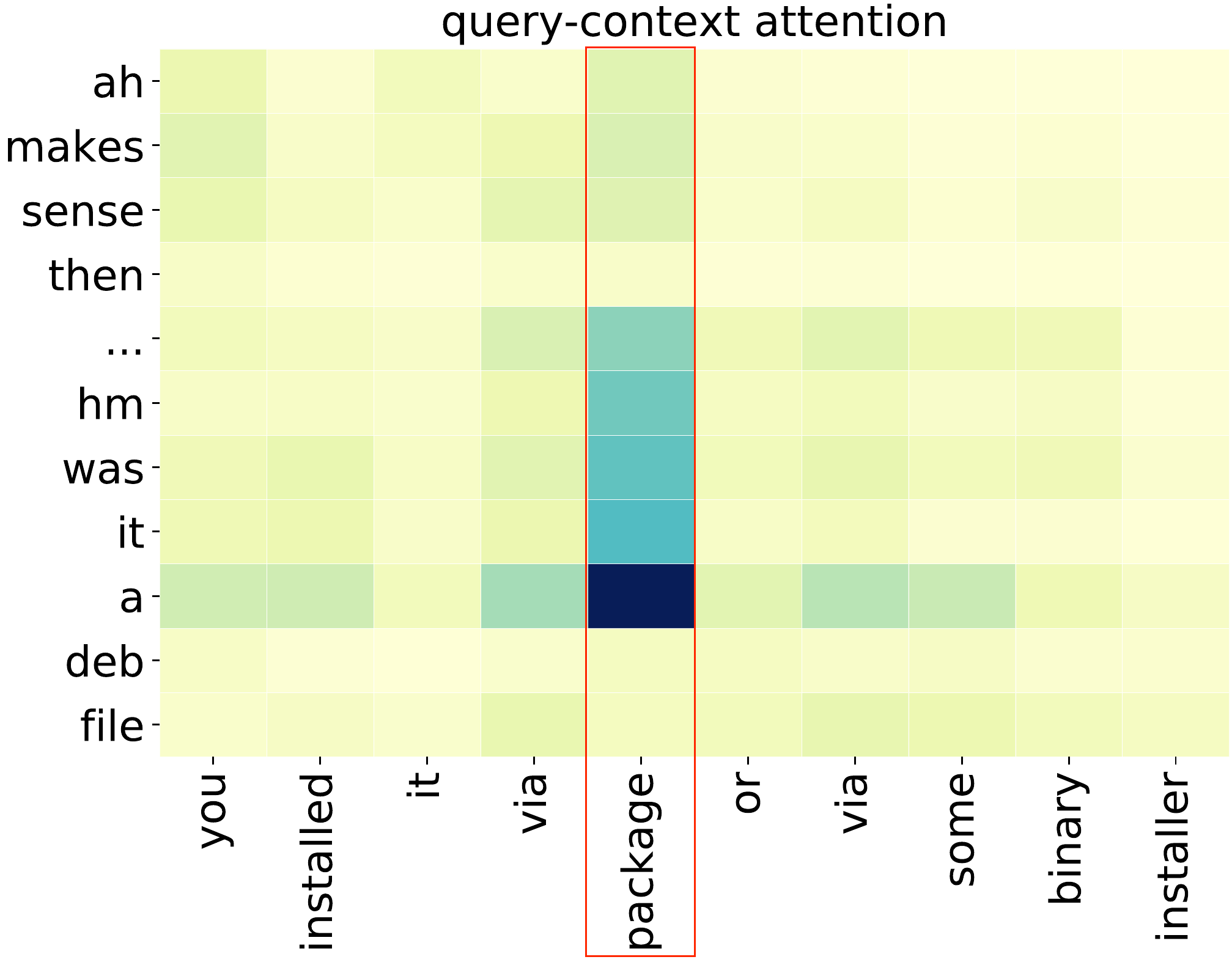}
\end{minipage}}
\subfigure{
\begin{minipage}[c]{0.31 \linewidth}
\includegraphics[width=1\linewidth]{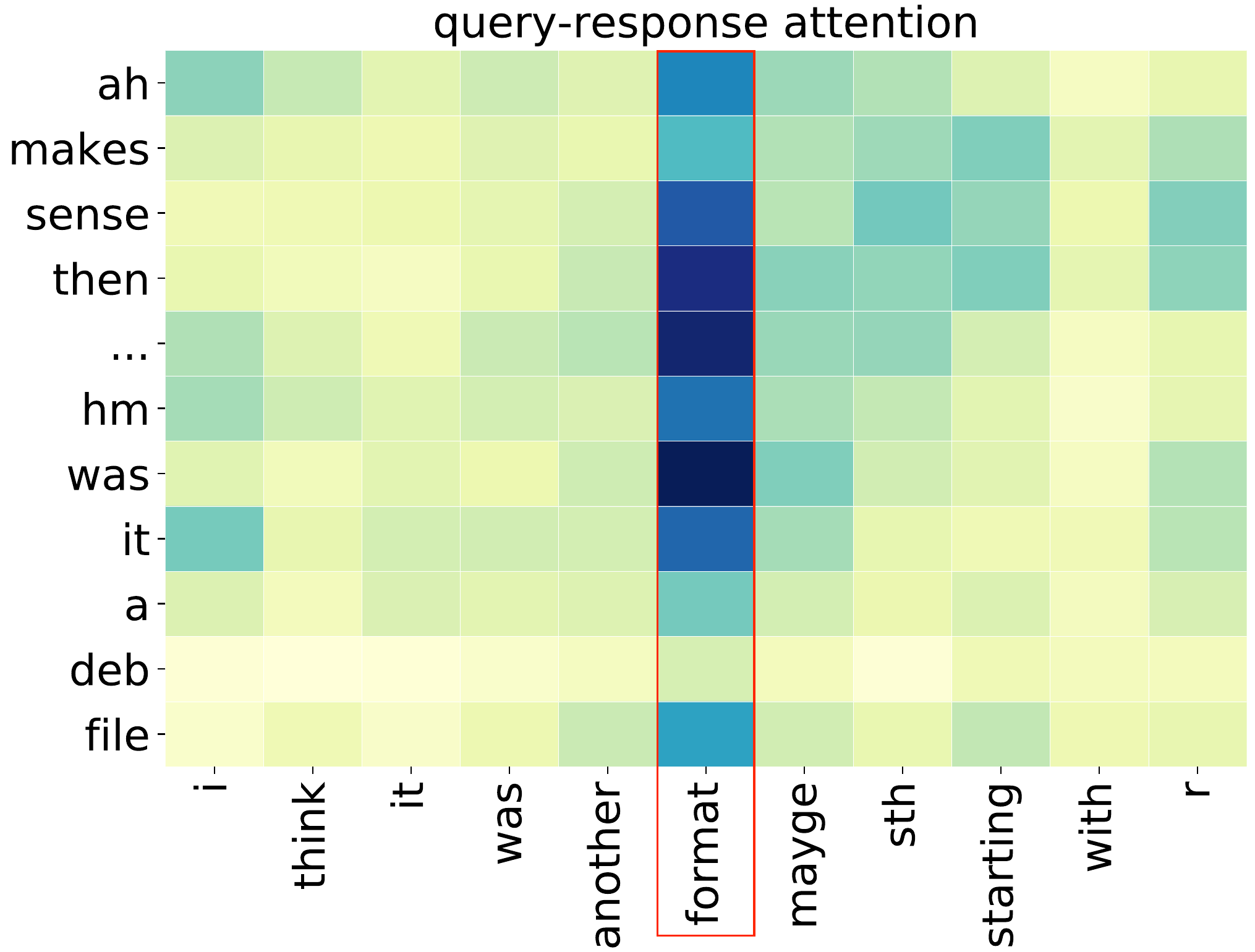}
\end{minipage}}
\subfigure{
\begin{minipage}[c]{0.31\linewidth}
\includegraphics[width=1\linewidth]{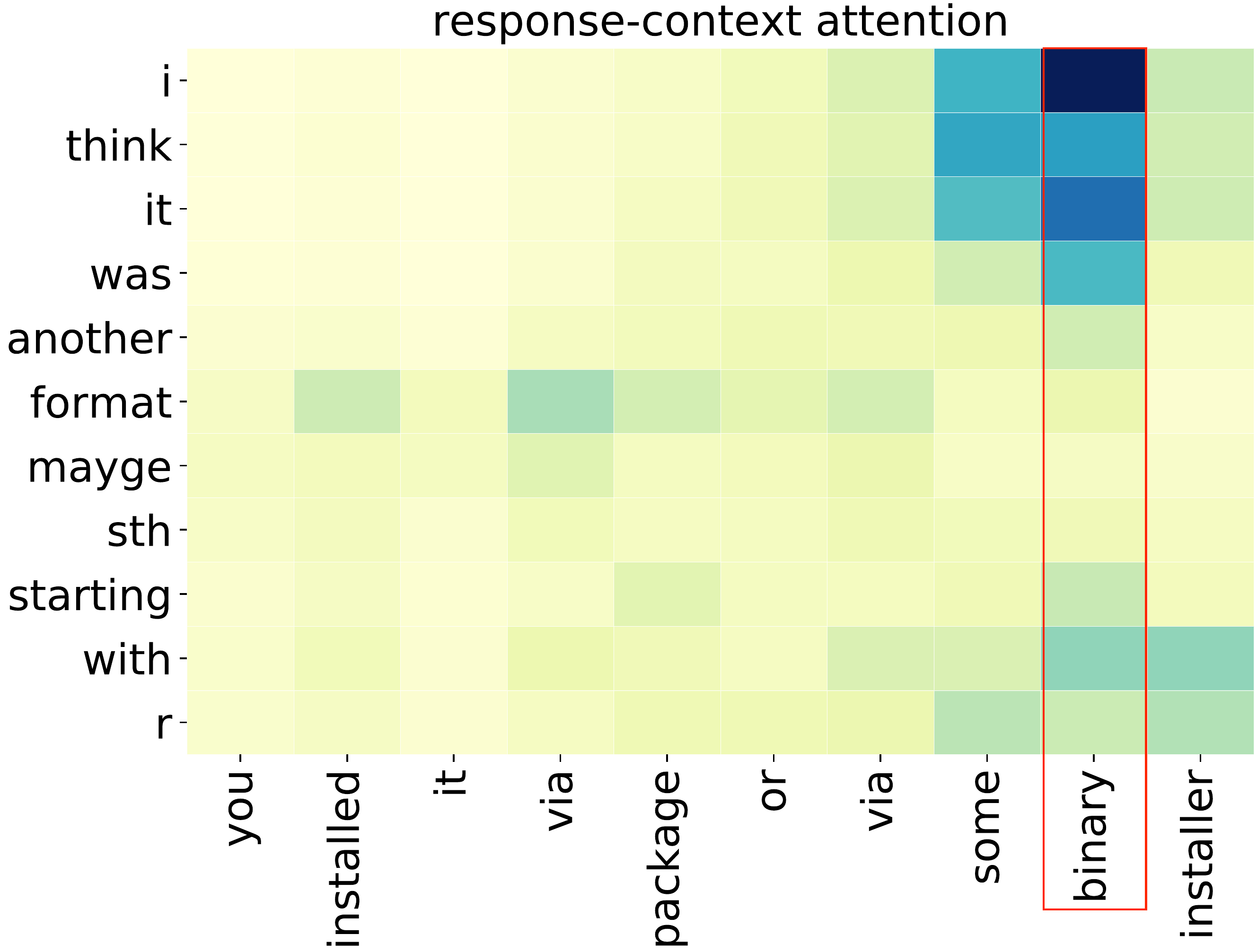}
\end{minipage}}
\caption{\label{bottom-up-att} The  attention visualization among the query, context, and response  in word-level. }
\end{figure*}

\subsection{Model Ablation}
To better demonstrate the effectiveness of TripleNet, we conduct the ablations on the model under the Ubuntu corpus for its larger data size.

We first remove the triple attention and matching parts (-TAM); the result shows a marked decline (2.4\% in $R_{10}$@1), which is in the second part of Table \ref{result-ablation}.
The performance of the model is similar to the baseline model DAM. This indicates that our four-level hierarchical representation may play a similar role to the five stacks Transformer in DAM.
We then remove the triple attention part, which means we match the pairs $\left\langle{C, R} \right\rangle$ and $\left\langle{Q, R} \right\rangle$ with their original representation in each level; the score of $R_{10}$@1 drops 1.4\%, which shows the effect of triple attention.
We also have tried to remove all the parts related to the query (-Query). That means the attention and matching parts are only calculated within the pair $\left\langle{C, R} \right\rangle$. 
It is worth mentioning that the information of the query is still contained at the end of the context.
The performance also has a marked drop (1.6\% in $R_{10}$@1), which shows that it is necessary to model the query separately.
To find out which subsection in those parts  is more important, we remove each one of them.

{\noindent \textbf {Triple attention matching ablation.} }
As we can see in the third part of Table \ref{result-ablation}, when attention between context and response is removed (-A$_{CR}$), the largest decrease (0.6\% in $R_{10}$@1) appears, which indicates that the relationship between context and response is most important in the triple. 
The attentions in the other two pairs $\left\langle{C, Q} \right\rangle$ and $\left\langle{Q, R} \right\rangle$ all lead to a slight performance drop (0.3 and 0.5 in $R_{10}$@1), which may be because they overlap with each other for updating the representation of the triple.

When we remove the matching between context and response, we find that the performance of the model has a marked drop (2.1 in $R_{10}$@1), which shows that the relationship within $\left\langle{C, R} \right\rangle$ is the base for selecting the response. 
The query and response matching part also leads to a significant decline. This shows that we should pay more attention to query within the whole context.

\begin {table} [tp]
\begin {center}\small
\begin {tabular} {l cccc}
\toprule
& R${_2}$@1& R$_{10}$@1& R$_{10}$@2 & R$_{10}$@5 \\
\midrule
TripleNet & 94.3 & 79.0 & 88.5 & 97.0\\
\midrule
~ -TAM & 93.5 & 76.6 & 86.8 & 96.6 \\
~ -A$_{tri}$ & 93.8 & 77.6 & 87.6 & 96.9 \\
~ -Query & 93.8 & 77.4 & 87.3 & 96.6 \\
\midrule
~ -A$_{CR}$ & 94.1 & 78.4 & 87.9 & 97.0 \\
~ -A$_{QR}$ & 94.1 & 78.5 & 88.1 & 97.0 \\
~ -A$_{CQ}$ & 94.3 & 78.7 & 88.3 & 97.0 \\
~ -M$_{CR}$ & 93.7 & 76.9 & 87.0 & 96.7 \\
~ -M$_{QR}$ & 94.4 & 78.5 & 88.1 & 97.1 \\
\midrule
~ -char & 94.1 & 78.3 & 88.0 & 97.1 \\
~ -word & 94.3 & 78.5 & 88.2 & 97.0 \\
~ -utterance & 94.1 & 78.6 & 88.1 & 97.1 \\
~ -context & 94.0 & 78.4 & 88.0 & 97.0 \\
\bottomrule
\end {tabular}
\end {center}
\caption {\label{result-ablation}Ablation studies on Ubuntu Dialogue Corpus. The letter `A' stands for the subsection in triple attention,  and `M' the is triple matching part.}
\end {table}

{\noindent \textbf {Hierarchical representation ablation.} }
To find out the calculation of which level is most important, we also tried to remove each level calculation from the hierarchical representation module, which can be seen in the fourth part of Table \ref{result-ablation}.
To our surprise, when we remove char (-char) and context level calculation (-context), we observe that the reduction (0.5 in $R_{10}$@1) is more significant than the other two, indicating that we should pay more attention to the lowest and highest level information.
Also by removing the other two levels, there is  also a significant reduction from TripleNet, which means each level of the three is indispensable for our TripleNet .

From the experiments in this part, we find that each subsection of the hierarchical representation module only leads to a slight performance drop. Maybe it's because the representation from each level represent the conversation from a unique and indispensable perspective,  and the information conveyed by different representations may have some overlap.

\section{Analysis and Discussion}\label{analysis}
\subsection{Visualization}
By decoding our model for the case in Figure \ref{case-ubuntu}, we find that our model TripleNet can choose the true response. 
To analyze in detail how triple attention works, we get the attention in word-level as the example and visualize it in Figure \ref{bottom-up-att}. 
As there are so many words in the context, we only use the second utterance in the upper part of Figure \ref{case-ubuntu} for its relatively rich semantics.

In the query-context attention, the query mainly pays attention to the keyword `package.' 
This is helpful to get the topic of the conversation.
While the attention of context focuses on the word `a' which is near the key phrase `deb file,' which may be because the representation of the word catches some information from the words nearby by Bi-LSTM.
In the query-response attention, the result shows that the attention of the query mainly focuses on the word  `format,' which is the most important word in the response. 
But we can also find that the response does not catch the important words in the query.
In the response-context attention, the response pays more attention to the word `binary,' which is another important word in the context.

From the three maps, we find that each attention can catch some important information but miss some useful information too.
 If we join the information in query-context and response-context attention, we can catch the most import information in the context. Furthermore, the query-response attention can help us catch the most important word in the response. 
 So it is natural for TripleNet to select the right response because the model can integrate the three attentions together.

\subsection {Discussion} 

\begin {figure} [t]
  \centering
  \includegraphics [width= 0.48\textwidth]  {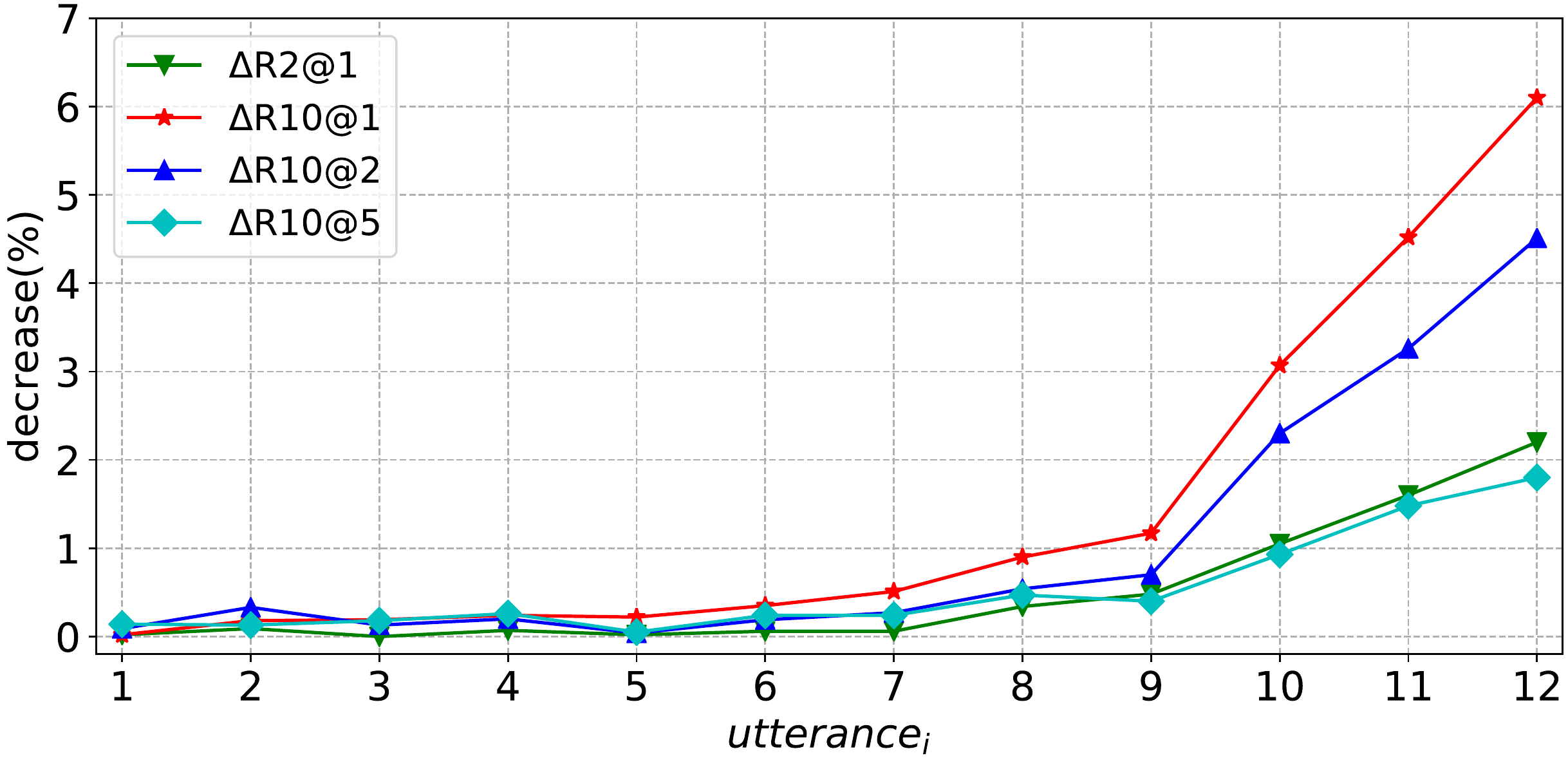}
  \caption{\label{result-query} The decrease of the performance when the $utterance_i$ is removed in Ubuntu Corpus.}
\end {figure}

 In this section, we will discuss the importance of different utterances in the context. To find out the importance of different utterances in the context, we conduct an experiment by removing each one of them with the model (-Query) in the ablation experiment part because the model deals all the utterances include the query in the same way.
For each experiment in this part, we remove the $i$th ($0<i<13$ and $Q=U_{12}$) utterance in the context both in training and evaluation processes and report the decrease of performance in  Figure  \ref {result-query}. 
We find that the removing of the query leads the most significant decline (more than 6\% in $R_{10}$@1), that indicates the query is much more important than any other utterances.
Furthermore, the decrease is stable before the $9$th  utterances and raises rapidly in the last 3 utterances.
We can deduce that the last three utterances are more important than the other ones.

From the whole result, we can conclude that it's better to model the query separately than deal all of the utterances in the same way for their significantly different importance; we also find that we should pay more attention to the utterances near the query because they are more important.
%%%%%%%%%%%%%%%%%%%%%%%%%%%%%%%%%%%%%%%%%
\section{Conclusion}\label{conclusion}
In this paper, we propose a model TripleNet for multi-turn  response selection.
We model the context from low (character) to high (context) level, update the representation by triple attention within $ \left\langle{C, Q, R} \right\rangle$, match the triple focused on response, and fuse the matching results with hierarchical LSTM for prediction.
Experimental results show that the proposed model achieves state-of-the-art results on both Ubuntu and Douban corpus, which ranges from a specific domain to open domain, and English to Chinese language, demonstrating the effectiveness and generalization of our model.
In the future, we will apply the proposed triple attention mechanism to other NLP tasks to further testify its extensibility. 
 
 \section*{Acknowledgement}\label{Acknowledgement}
 We would like to thank all anonymous reviewers for their hard work on reviewing and providing valuable comments on our paper. We also would like to thank Yunyi Anderson for proofreading our paper thoroughly. This work is supported by National Key R\&D Program of China via grant 2018YFC0832100.

\bibliography{conll-2019}
\bibliographystyle{acl_natbib}
\end{document}